\documentclass[conference]{IEEEtran}
\IEEEoverridecommandlockouts
\usepackage[bookmarks=false,pdfpagemode=UseNone]{hyperref}
\usepackage{cite}
\usepackage{amsmath,amssymb,amsfonts}
\usepackage[ruled,vlined]{algorithm2e}
\SetKwFunction{KwCall}{call}
\usepackage{booktabs}
\usepackage{mathtools}
\usepackage{graphicx}
\usepackage{subcaption}
\usepackage{textcomp}
\usepackage[dvipsnames]{xcolor}
\usepackage{pgfplots}
\pgfplotsset{compat=newest}
\usepackage{mathrsfs}
\usepackage{multirow}
\usepackage{booktabs}
\usepackage{makecell}
\usetikzlibrary{arrows}
\usetikzlibrary{arrows.meta, backgrounds, calc, shapes.geometric, shapes.symbols, decorations.pathreplacing}
\def\BibTeX{{\rm B\kern-.05em{\sc i\kern-.025em b}\kern-.08em
    T\kern-.1667em\lower.7ex\hbox{E}\kern-.125emX}}
    
\begin{document}
\title{\textit{Chisme}: Heterogeneity-Aware Gossip Learning}
\author{\IEEEauthorblockN{Harikrishna Kuttivelil}
\IEEEauthorblockA{\textit{Computer Science \& Engineering} \\
\textit{University of California}\\
Santa Cruz, California, USA \\
hkuttive@ucsc.edu}
\and
\IEEEauthorblockN{Katia Obraczka}
\IEEEauthorblockA{\textit{Computer Science \& Engineering} \\
\textit{University of California}\\
Santa Cruz, California, USA \\
katia@soe.ucsc.edu}
}
\maketitle
\begin{abstract}
As end-user device capability increases and the amount of data at the Internet's edge rises, distributed learning has emerged as a key enabling technology for the intelligent edge.
Existing approaches like federated learning (FL) and decentralized FL (DFL) enable privacy-preserving distributed learning among clients, while gossip learning (GL) approaches have emerged to address the potential challenges in resource-constrained, connectivity-challenged infrastructure-less environments.
However, most distributed learning approaches assume largely homogeneous data distributions and may not consider or exploit the heterogeneity of clients and their underlying data distributions.
This paper introduces \textit{Chisme}, a novel fully decentralized distributed learning algorithm designed to address the challenges of implementing robust intelligence in network edge contexts characterized by heterogeneous data distributions, episodic connectivity, and sparse network infrastructure or lack thereof.
\textit{Chisme} leverages the inferred affinity between clients' underlying data distributions calculated from received model exchanges to inform how much influence received models have when merging into the local model.
By doing so, it enables clients to strategically balance between broader collaboration to build more general knowledge and more selective collaboration to build specific knowledge.
We evaluate \textit{Chisme} against contemporary approaches using image recognition and time-series prediction scenarios while considering different network connectivity conditions, representative of real-world distributed intelligent systems running at the network's edge.
Our experiments demonstrate that \textit{Chisme} outperforms other contemporary distributed learning approaches in almost every case -- clients using \textit{Chisme} exhibit faster training convergence, lower final loss after training, and lower performance disparity between clients.
\end{abstract}
\begin{IEEEkeywords}
federated learning, distributed deep learning, decentralized deep learning, personalized federated learning, internet of things
\end{IEEEkeywords}
\section{Introduction}
As network-connected devices become more pervasive and the amount of data generated at the network edge rises significantly, utilizing its potential to enrich services and applications for its users while contending with a distributed and privacy-focused landscape is an increasingly pertinent problem.
As edge hardware and algorithms have become more capable and efficient, data-driven artificial intelligence and machine learning (AI/ML) operations have become increasingly ubiquitous on the network edge, manifesting as a more distributed and democratized intelligence landscape and empowering numerous domains, e.g. smart cities, healthcare, and autonomous vehicles~\cite{zhou2019edge}.
While federated learning (FL)~\cite{mcmahan2017federated} has emerged as a foundational distributed training paradigm, existing FL-based approaches face limitations: (1) some can be ineffective in volatile networks due to dependence on centralized infrastructure; and (2) most jointly train a global, \textit{one-size-fits-all} model across all clients, which may sufficiently serve applications with homogeneous data distributions -- i.e., distributions in which each client's data are broadly similar in nature and characteristics -- but not for applications with more heterogeneous data distributions.

In this paper, we introduce \textit{Chisme} (Collaborative Heterogeneity-aware Intelligence via Similarity-based Model Exchange), a novel fully decentralized gossip learning (GL) approach that facilitates collaborative training between clients of models tailored towards their underlying data distributions.
Our work contributes the following --
(1) we derive a \textit{data affinity} heuristic by leveraging the cosine similarity between local model progressions and asynchronously received model parameters;
(2) we extend standard gossip learning by leveraging that inferred affinity to inform the influence of received model parameters as they are merged into the local model; and
(3) we evaluate \textit{Chisme} against related approaches across multiple application and network scenarios, demonstrating its effectiveness in achieving better accuracy, lower loss, faster convergence, and less performance disparity between clients.
\section{Related Work}\label{sec:bg-related}

Traditional deep learning has historically required more computing and more data than what was available on a typical edge device, necessitating these devices to transmit large amounts of encrypted data over the network to be collected and processed at data centers, making these systems vulnerable to high communication costs, network connectivity challenges, security risks, and privacy concerns.
As devices became more computationally capable, federated learning (FL)~\cite{mcmahan2017federated} was proposed as a solution to leverage on-device training, in which a central server facilitates model training over data distributed across many clients, where clients' data never leave their devices.
However, while FL has been quickly adopted as a core enabling paradigm for distributed application domains, it has inherent limitations.
One limitation is that standard FL depends on a server to facilitate the dissemination and aggregation of model updates, with performance hinging on server and network availability.
To address these limitations, decentralized FL (DFL)~\cite{lalitha2018fully, liu2022decentralized} and gossip learning (GL)~\cite{hegedHus2019gossip}-- in which agents within a network act as both the server and the client -- have been offered as alternatives to FL to address the limitations posed by centralization.
DFL typically refers to approaches in which agents carry out their local training, communication, and aggregation in synchronous phases as a network;
GL, on the other hand, typically refers to approaches where communication is more asynchronous and sparse, with agents processing and merging received models into their own as they come in instead of waiting to enter an aggregation phase.
The other limitation in FL, including its decentralized variants, is the global objective to train a \textit{one-size-fits-all} model, adequate for scenarios with more homogeneous data distributions but inadequate for those with more heterogeneous or incongruent data distributions, where multiple optimal solutions exist across the network, and a compromised solution that may not be sufficient for all clients~\cite{sattler2020cfl}.

This motivates the need for more \textit{heterogeneity-aware} approaches, i.e., approaches that facilitate collaborative model training while allowing multiple optimal solutions across clients or distributions.
To this end, several approaches have leveraged global and local model training to yield heterogeneity-aware differentiated solutions, including hierarchical FL approaches~\cite{liu2020clienthierarchical} and multi-phase approaches with both federated and on-device training~\cite{huang2021personalized, arivazhagan2019silopersonalized}.
However, these approaches largely do not consider the similarity of underlying data distributions, and therefore cannot efficiently capitalize on collaboration between compatible clients.
Clustered FL~\cite{sattler2020cfl, castellon2022flic} has been proposed to infer similarity from model updates to structure federated learning within incrementally partitioned groups of clients with similar data distributions.
Still, they rely on a central server to orchestrate heterogeneity-aware federated learning, whereas our proposed approach does not.
Decentralized approaches that target the personalization of models are part of a much more recent paradigm in research.
Ma et al.~\cite{ma2022like} use the cosine similarity of model parameters to optimize regularization and loss functions towards personalization, while Zec et al.~\cite{zec2024effects} consider cosine similarity and other similarity measures for model aggregation.
However, these DFL approaches require synchronous communication and aggregation phases that may make them susceptible to delays caused by straggling devices or require synchronous control planes, whereas our approach leverages gossip learning to address these risks.
To our knowledge, \textit{Chisme} is the first heterogeneity-aware gossip learning protocol that leverages data affinity inferred from exchanged model parameters to influence model merging to better serve clients in a network of heterogeneous data distributions, while maintaining model robustness through collaboration.

\section{Chisme} \label{sec:chisme}

\textit{Chisme} leverages cosine similarity between GL model exchanges to infer affinity between participating clients' underlying data distributions and applies it to influence model merging to provide an effective, efficient, and equitable client models that balance between broader and more selective collaboration.
Table~\ref{tab:symbols} summarizes all variables and symbols used throughout this section.
\begin{table}[b]
  \centering
  \caption{Symbol Reference}
  \label{tab:symbols}
  \begin{tabular}{c|l}
    \textsc{symbol} & \textsc{description} \\
    \hline
    $N$ & the network \\ 
    $\mathcal{C}_N$ & node connectivity rate of $N$ \\
    $\mathcal{R}_N$ & network loss rate of $N$ \\
    $M_x$ & neighbors of $x$; $\subset N$ \\
    $R_x^t$ & neighbors of $x$ connected to $x$ at $t$; $\subset M_x$ \\
    $m_x^t$ & message from $x$ at $t$\\ \hline
    $T$ & all time containing all time-steps \\ 
    $t_{(x)}$ & specified round/time-step (at $x$) \\ 
    $*$ & arbitrary time or entity of remote origin \\ \hline
    $\theta_x^{(t)}$ & model params. at $x$ (at $t$) \\ 
    $\beta$ & number of epochs per round \\ 
    $\mathcal{D}_x$ & dataset at $x$ \\ 
    $\mathcal{P}_x$ & underlying distribution containing $\mathcal{D}_x$ \\
    $\mu_x^{(t)}$ & model training experience of $x$ (at $t$) \\
    $\mathcal{M}_x$ & mapping of recent $\mu_k$ received at $x$ (Eq.~\ref{eq:exp-map}) \\ \hline
    $|z|$ & size of structure $z$ \\
    $\Delta_{x,y}$ & difference vector of params. of $x$, $y$ \\
    $S(a,b)$ & cosine similarity of $a$, $b$ \\ \hline
    $\tau_x$ & timestep of last saved model state at $x$ \\ 
    $\alpha$ & experience-based influence (Eq.~\ref{eq:base-gl})\\
    $S'(a, b)$ & scaled cosine similarity of $a$, $b$ (Eq.~\ref{eq:scaled-cos-sim})\\
    $\omega$ & similarity heuristic value (Eq.~~\ref{eq:chisme-gl-sim})\\
    $\eta$ & similarity heuristic-adjusted influence (Eq.~\ref{eq:chisme-gl-coeff})\\
  \end{tabular}
\end{table}

\paragraph*{System Model}\label{sec:system-model}
We consider a decentralized network of clients $N$, where every client $x \in N$ has a local dataset $\mathcal{D}_x$ from its local distribution $\mathcal{P}_x$ and is instantiated with the same baseline model $\theta^{t=0}$.
We assume that each client $x$ periodically trains and updates its local model parameters $\theta_x$ on its local dataset $\mathcal{D}_x$.
We consider heterogeneous data distributions such that for any clients $i,j \subset N$ where $i \neq j$, it is possible that their distributions $\mathcal{P}_i, \mathcal{P}_j$ may not be similar.
We assume that each client $x$ is connected to its neighbors $M_x \subset N$ by an underlying network topology that remains static for the duration of the system runtime.
We assume that the links along this topology can be volatile, such that at a given timestep $t$, client $x$ may only be able to communicate model update messages $m_x^t$ to a subset of its neighbors $R_x^t \subseteq M_x$.
Given this system model, we propose to address the following problem ---
how can clients $i \in N$ collaborate in training their models $\theta_i$ on their local data distributions $\mathcal{P}_i$ without relying on centralized infrastructure and using only the communication of model update messages $m$ between clients to achieve lower loss and faster convergence?

\subsection{Preliminary on Gossip Learning}\label{sec:gl}

Consider the FL-based distributed training approach in which client $i \in N$ is initialized with a local model with identical model parameters $\theta$, where periodically at time $t$, clients train their local model $\theta_i^t$ for $\beta$ epochs on their local dataset $\mathcal{D}_i$, and send a message $m_i^t$ containing their updated model parameters $\hat\theta_i^t$ and data size $|\mathcal{D}_i|$ to their neighbors $M_i \subset N$.
In standard FL and DFL, the training, communication, and aggregation of models occur \textit{synchronously}, such that each round involves an interval in which all clients train followed by an interval where all clients $i$ transmit their updated model parameters $\hat\theta_i^t$ to a server (in centralized FL) or other clients $k\ \forall \ k \in R_x^t$ (in DFL), after which the round's updates are aggregated using weighted averaging~\cite{mcmahan2017federated}.
However, in gossip learning, where the training, communication, and aggregation of models can occur asynchronously, each client $i$ merges each received update's parameters $\theta_k^*$ as it is received from connecting neighbors $k \in R_i^{t_i}$ into its own local model $\theta_i$.
Because it is possible that at any given time of exchange $t^* \in T$ between clients, the total number of training iterations $\beta_x$ for each of the exchanging clients $x$ may be different, instead of communicating data size as done in many FL and DFL approaches, clients communicate their model training \textit{experience} $\mu_x^t = t \times \beta \times \allowdisplaybreaks|\mathcal{D}_x|$ following the baseline GL implementation\cite{hegedHus2019gossip}.
Thus, when a client $i$ merges a model update from another client $k$ into its own, the mixing weight of the incoming model is based on \textit{merge influence} $\alpha$ calculated using the received update's experience $\mu_k^*$ and the client's local experience $\mu_i$:
\begin{equation}\label{eq:base-gl}
    \begin{split}
        \alpha &= \frac{\mu_k^*}{\mu_i + \mu_k^*} \\
        \theta_i &\leftarrow (1-\alpha)\ \theta_i + \alpha\theta_k^* \\
        \mu_i &\leftarrow \mathrm{max}(\mu_i, \mu_k^*)
    \end{split}
\end{equation}
While this baseline GL merge method (Eq.~\ref{eq:base-gl}) works well for homogeneous scenarios (where $P_i \sim P_k \ \forall \ k\in N$), in heterogeneous scenarios (where $P_i \sim P_k \ \neg\forall \ k\in N$), this may result in gradient conflict, training volatility, and ultimately slower convergence to optimal solutions.
Therefore, in our approach, we extend GL such that each client $i$ retains the most recent model experience -- including its own after either training or merging -- values per client $k \in M_i$ in local map $\mathcal{M}_i$:
\begin{equation} \label{eq:exp-map}
    \begin{split}
        \mathcal{M}_i^i \leftarrow \mu_i \ \text{(after training or merge)}\\
        \mathcal{M}_i^k \leftarrow \mu_k \ \text{(upon reception of $m_k^{*}$)}
    \end{split}
\end{equation}
Then, the influence $\alpha_k$ of a received model update $\theta_k$ is calculated using the received update's experience $\mu_k^*$ and all prior merged or locally trained experiences $\mathcal{M}_i$, whereas the local experience is calculated with the influence:
\begin{equation}\label{eq:mod-gl}
    \begin{split}
        \alpha_k &= \frac{\mu_k^*}{\sum_j^{\mathcal{M}} \mu_j} \\
        \theta_i &\leftarrow (1-\alpha_k)\ \theta_i + \alpha_k \theta_k^* \\
        \mu_i &\leftarrow (1-\alpha_k)\ \mu_i + \alpha_k \mu_k^*
    \end{split}
\end{equation}

\subsection{Inferring Data Similarity through Cosine Similarity} \label{sec:cos-sim}
Given our network of clients $N$, we know an optimal solution for $\theta_i$ exists for each client $i$'s local data distribution $\mathcal{P}_i$.
As a client trains its model parameters, each gradient/weights update moves the client's model toward that optimal solution, and thus two similar clients' models would train towards similar optimal solutions and will have low angular distance and two dissimilar clients' models would diverge towards dissimilar optimal solutions and will have a high angular distance.
We leverage cosine similarity to measure this angular distance between model progression vectors to infer data distribution affinity, which has proven its efficiency in existing approaches~\cite{sattler2020cfl, castellon2022flic, ma2022like, zec2024effects}.
However, these approaches rely on model parameter deltas calculated from a common baseline state or an intermediary state achieved via synchronous model exchanges, making it unsuitable for the gossip learning context.
Therefore, to infer the affinity of the underlying data distributions between a local client $i$ and neighboring client $k$ at some time $\tau \in T$, we calculate the deltas $(\Delta_{i,i},\Delta_{k,i})$ of the model parameters $(\theta_i^{\tau}, \theta_k^{\tau})$ w.r.t. the local prior model parameter state $\theta_i^{t<\tau}$ as shown in Eq.~\ref{eq:update-deltas}, and use the cosine similarity $S$ of the deltas as shown in Eq.~\ref{eq:cos-sim}.
\begin{equation} \label{eq:update-deltas}
    \begin{split}
        \Delta_{i,i} &= \theta_i^{\tau} - \theta_i^{t<\tau} \\
        \Delta_{k,i} &= \theta_k^{\tau} - \theta_i^{t<\tau}
    \end{split}
\end{equation}
\begin{equation} \label{eq:cos-sim}
    S(\Delta_{i,i}, \Delta_{k,i}) = \frac{\Delta_{i,i} \cdot \Delta_{k,i}}{||\Delta_{i,i}|| ||\Delta_{k,i}||} \in [-1, 1]
\end{equation}
We will use this inferred data affinity to scale the contribution of received model updates, therefore we scale cosine similarity to be non-negative (Eq.~\ref{eq:scaled-cos-sim}).
\begin{equation}\label{eq:scaled-cos-sim}
    S'(\Delta_{i,i}, \Delta_{k,i}) = \frac{S(\Delta_{i,i}, \Delta_{k,i}) + 1}{2} \in [0, 1]
\end{equation}





\subsection{Fully Decentralized Heterogeneity-Aware Deep Learning}\label{sec:chisme-main}

In \textit{Chisme}, we utilize the cosine similarity-based data affinity heuristic (Section~\ref{sec:cos-sim}) in gossip learning's model merging mechanism to create a robust and accelerated heterogeneity-aware decentralized model training approach. 
We describe below the processes that occur independently at each client $i$ in the collaborative decentralized network $N$.

\begin{enumerate}
    \item \textsc{initialization:} In network $N$, each client $i \in N$ is initialized with initial model parameters $\theta^{t=0}$. The model training experience is initialized to zero.
    \begin{equation}
        \begin{split}\label{eq:chisme-init}
            \theta_i^{t_i=0} &\leftarrow \theta^{t=0} \\
            \mu_i^{t_i=0} &\leftarrow 0 \\
            \forall\ i &\in N
        \end{split}
    \end{equation}

    \item \textsc{training:} At any time, each client $i \in N$ snapshots its local model parameters to $\theta_i^{\tau_i}$ and trains its local model on its local dataset $\mathcal{D}_i$ to procure updated model parameters $\theta_i^{t_i+\zeta}$.
    It also updates its model experience $\mu_i^{t_i}$ to reflect the number of samples it has trained on.
    \begin{equation}
        \begin{split}\label{eq:checkpoint-train}
            \theta_i^{\tau_i} &\leftarrow \theta_i^{t_i} \\
            \theta_i^{t_i} &\leftarrow \mathrm{train}(\theta_i^{t_i}, \mathcal{D}_i)
        \end{split}
    \end{equation}
    \begin{equation}
        \begin{split}\label{eq:chisme-update-exp}
            \mu_i^{t_i} \leftarrow \mu_i^{t_i} + |\mathcal{D}_i| \\
            \mathcal{M}_i^i \leftarrow \mu_i^{t_i}
        \end{split}
    \end{equation}

    \item \textsc{communication:} After each client $i \in N$ trains, it creates messages $m_i^{t_i}$ containing its updated model parameters $\theta_i^{t_i}$ and its updated experience $\mu_i^{t_i}$ and transmits it to connected clients $M_i \subseteq N$. Note that we continue to refer to client $i$'s current local timestep as $t_i$.
    \begin{equation} \label{eq:chisme-package}
            m_i^{t_i} \leftarrow \left\{\theta_i^{t_i}, \mu_i^{t_i}\right\}
    \end{equation}
    At any time, the client may receive messages $m_k^*$ from some client $k \in R_i^{t_i}$ where $R_i^{t_i} \subseteq M_i$ given the possibility of lossy connections.

    \item \textsc{analysis:} When a client $i$ receives a message $m_k^*$, it stores the experience $\mu_k^*$ in local map $\mathcal{M}_i$ derives the experience-based influence $\alpha_k$ as in Eq.~\ref{eq:mod-gl}.  
    \begin{equation}\label{eq:chisme-exp-inf}
        \begin{split}
            \theta_k^*, \mu_k^* &\leftarrow m_k* \\
            \mathcal{M}_i^k &\leftarrow \mu_k^* \\
            \alpha_k &= \frac{\mu_k^*}{\sum_j^{\mathcal{M}} \mu_j}
        \end{split}
    \end{equation}
    Then, client $i$ calculates the update deltas w.r.t. saved local prior state $\theta_i^{\tau_i}$ as in Eq.~\ref{eq:update-deltas} and then calculates the scaled cosine similarity as in Eq.~\ref{eq:scaled-cos-sim} normalized against the local identity similarity of $1$, $\omega$.
    \begin{equation*}
        \begin{split}
            \Delta_{i,i} &= \theta_i^{t_i} - \theta_i^{\tau_i} \\
            \Delta_{k,i} &= \theta_k^* - \theta_i^{\tau_i} \\
        \end{split}
    \end{equation*}
    \begin{equation}\label{eq:chisme-gl-sim}
        \omega_{i,k} = \frac{S'(\Delta_{i,i}, \Delta_{k,i})}{1 + S'(\Delta_{i,i}, \Delta_{k,i})}
    \end{equation}
    Finally, we use the experience-based influence $\alpha_k$ and the similarity heuristic $\omega_{i,k}$ to calculate the combined influence coefficient $\eta_k$.
    \begin{equation} \label{eq:chisme-gl-coeff}
        \eta_k = \frac{\alpha_k\omega_{i,k}}{(1-\alpha_k)(1-\omega_{i,k})+\alpha_k\omega_{i,k}}
    \end{equation}

    \item \textsc{merge:} Each client $i$ uses the combined influence weight coefficient $\eta^*$ to merge the received model parameters into its local model, similar to Eqs.~\ref{eq:base-gl} and \ref{eq:mod-gl}.
    \begin{equation} \label{eq:chisme-gl-merge}
        \theta^{t_i}_i \leftarrow (1-\eta_k)\ \theta_i^{t_i} + \eta_k\theta_k^*
    \end{equation}
    The local model experience also uses the same $\eta_k$ to update its model experience to represent the weighted averaging of the merged models and the local map $\mathcal{M}_i$ is updated.
    \begin{equation} \label{eq:chisme-gl-exp}
        \begin{split}
            \mu_i &\leftarrow (1-\eta^*)\ \mu_i^{\tau_i + \zeta} + \eta^*\mu_k^* \\
            \mathcal{M}_i^i &\leftarrow \mu_i
        \end{split}
    \end{equation}
\end{enumerate}

\paragraph*{Note on Memory \& Convergence}
Our method has each client store its local active model parameters $\theta^{t}$, its saved prior model parameters $\theta^\tau$, and the received model parameters $\theta^*$ that are being analyzed and merged during an interaction.
Given a constant number of parameters per model $p = |\theta|$, its memory usage can be described as $3p$, which stays constant even as the network size increases.
In the case where clients may receive messages it can not yet process, it may consider storing these messages in a fixed-size $\beta$ buffer; then memory usage would bes $(3 + \beta)p$.
We consider the memory footprint of $\mathcal{M}$ to be trivial given that it contains integers and $|N|^2 \ll p$.
Because \textit{Chisme} actuates model aggregation similarly to standard GL~\cite{hegedHus2019gossip} and FL~\cite{mcmahan2017federated, sattler2020cfl, castellon2022flic} approaches that are known to converge when collaboration is strong and mimics classical on-device learning when collaboration is low, we can assume that it will converge within these bounded convergences.
We validate this convergence empirically in our experiments.
\section{Experiments} \label{sec:results}
To demonstrate the effectiveness of \textit{Chisme} against other collaborative learning approaches, we use image recognition and weather prediction scenarios as proxies for real-world network edge and IoT use cases.
\newcommand{\progRounds}{}
\newcommand{\progMetric}{}
\newcommand{\progMeanFile}{}
\newcommand{\finalMetric}{}
\newcommand{\finalMeanFile}{}
\newcommand{\finalPTenFile}{}
\newcommand{\finalPNinetyFile}{}
\newcommand{\buildGraph}[7] 
{
\renewcommand{\progRounds}{#1}
\renewcommand{\progMetric}{#2}
\renewcommand{\progMeanFile}{#3-mean.csv}
\renewcommand{\finalMetric}{#4}
\renewcommand{\finalMeanFile}{#5-mean.csv}
\renewcommand{\finalPTenFile}{#5-p10.csv}
\renewcommand{\finalPNinetyFile}{#5-p90.csv}
\begin{subfigure}[b]{0.48\textwidth}
    \centering
    \begin{tikzpicture}
        \begin{axis}[
            title={\textsc{Mean Client \progMetric\ Progression}},
            title style={at={(0.5,-0.2)}, anchor=north},
            font=\scriptsize,
            ylabel={\textsc{\progMetric}},
            xmin=0, xmax=\progRounds,
            xticklabel={\ifdim\tick pt=0pt \textsc{rounds}\else\pgfmathprintnumber{\tick}\fi},
            width=\linewidth,
            height=4cm,
            grid=major,
        ]
            \addplot[color=cyan, mark=none] table [x expr=\coordindex + 1, y=FederatedLearning, col sep=comma, header=true] {\progMeanFile};
            \addplot[color=blue, mark=none] table [x expr=\coordindex + 1, y=DecentralizedFederatedLearning, col sep=comma, header=true] {\progMeanFile};
            \addplot[color=teal, mark=none] table [x expr=\coordindex + 1, y=GossipLearning, col sep=comma, header=true] {\progMeanFile};
            \addplot[color=violet, mark=none] table [x expr=\coordindex + 1, y=FLIncrementalClustering, col sep=comma, header=true] {\progMeanFile};
            \addplot[color=orange, mark=none] table [x expr=\coordindex + 1, y=ChismeLearningBalanced, col sep=comma, header=true] {\progMeanFile};
        \end{axis}
    \end{tikzpicture}
    \pgfplotstableread[col sep=comma, header=true]{\finalMeanFile}\meantable
    \pgfplotstableread[col sep=comma, header=true]{\finalPTenFile}\ptentable
    \pgfplotstableread[col sep=comma, header=true]{\finalPNinetyFile}\pninetytable
    \pgfplotstablegetrowsof{\meantable}
    \pgfmathtruncatemacro{\lastrow}{\pgfplotsretval - 1}
    \pgfplotstablegetelem{\lastrow}{FederatedLearning}\of{\meantable}               \pgfmathsetmacro{\meanFL}{\pgfplotsretval}
    \pgfplotstablegetelem{\lastrow}{DecentralizedFederatedLearning}\of{\meantable}  \pgfmathsetmacro{\meanDFL}{\pgfplotsretval}
    \pgfplotstablegetelem{\lastrow}{GossipLearning}\of{\meantable}                  \pgfmathsetmacro{\meanGL}{\pgfplotsretval}
    \pgfplotstablegetelem{\lastrow}{FLIncrementalClustering}\of{\meantable}         \pgfmathsetmacro{\meanFLIC}{\pgfplotsretval}
    \pgfplotstablegetelem{\lastrow}{ChismeLearningBalanced}\of{\meantable}          \pgfmathsetmacro{\meanChisme}{\pgfplotsretval}
    \pgfplotstablegetelem{\lastrow}{FederatedLearning}\of{\ptentable}                \pgfmathsetmacro{\ptenFL}{\pgfplotsretval}
    \pgfplotstablegetelem{\lastrow}{DecentralizedFederatedLearning}\of{\ptentable}   \pgfmathsetmacro{\ptenDFL}{\pgfplotsretval}
    \pgfplotstablegetelem{\lastrow}{GossipLearning}\of{\ptentable}                   \pgfmathsetmacro{\ptenGL}{\pgfplotsretval}
    \pgfplotstablegetelem{\lastrow}{FLIncrementalClustering}\of{\ptentable}          \pgfmathsetmacro{\ptenFLIC}{\pgfplotsretval}
    \pgfplotstablegetelem{\lastrow}{ChismeLearningBalanced}\of{\ptentable}           \pgfmathsetmacro{\ptenChisme}{\pgfplotsretval}
    \pgfplotstablegetelem{\lastrow}{FederatedLearning}\of{\pninetytable}                \pgfmathsetmacro{\pninetyFL}{\pgfplotsretval}
    \pgfplotstablegetelem{\lastrow}{DecentralizedFederatedLearning}\of{\pninetytable}   \pgfmathsetmacro{\pninetyDFL}{\pgfplotsretval}
    \pgfplotstablegetelem{\lastrow}{GossipLearning}\of{\pninetytable}                   \pgfmathsetmacro{\pninetyGL}{\pgfplotsretval}
    \pgfplotstablegetelem{\lastrow}{FLIncrementalClustering}\of{\pninetytable}          \pgfmathsetmacro{\pninetyFLIC}{\pgfplotsretval}
    \pgfplotstablegetelem{\lastrow}{ChismeLearningBalanced}\of{\pninetytable}           \pgfmathsetmacro{\pninetyChisme}{\pgfplotsretval}
    \begin{tikzpicture}
        \begin{axis}[
            title={\textsc{Final \finalMetric : mean, $P_{10}$--$P_{90}$}},
            title style={at={(0.5,-0.35)}, anchor=north},
            font=\scriptsize,
            width=\linewidth,
            height=3cm,
            xbar,
            bar width=6pt,
            bar shift=0pt,
            ytick={1,2,3,4,5},
            ytick style={draw=none},
            yticklabels={Chisme, FLIC, GL, DFL, FL},
            y tick label style={font=\scriptsize},
            ymin=0.4, ymax=5.6,
            grid=major,
            grid style={line width=0.2pt},
            ymajorgrids=false,
            enlarge x limits={lower=false},
            error bars/x dir=both,
            error bars/x explicit,
            error bars/error bar style={black, line width=0.8pt},
            error bars/error mark options={black, rotate=90, mark size=3pt, line width=0.8pt},
        ]
            \addplot[fill=cyan,   draw=cyan]   coordinates {(\meanFL,    5) -= (\meanFL     - \ptenFL,    0) += (\pninetyFL     - \meanFL,    0)};
            \addplot[fill=blue,   draw=blue]   coordinates {(\meanDFL,   4) -= (\meanDFL    - \ptenDFL,   0) += (\pninetyDFL    - \meanDFL,   0)};
            \addplot[fill=teal,   draw=teal]   coordinates {(\meanGL,    3) -= (\meanGL     - \ptenGL,    0) += (\pninetyGL     - \meanGL,    0)};
            \addplot[fill=violet, draw=violet] coordinates {(\meanFLIC,  2) -= (\meanFLIC   - \ptenFLIC,  0) += (\pninetyFLIC   - \meanFLIC,  0)};
            \addplot[fill=orange, draw=orange] coordinates {(\meanChisme,1) -= (\meanChisme - \ptenChisme,0) += (\pninetyChisme - \meanChisme,0)};
        \end{axis}
    \end{tikzpicture}
    \caption{#6}
    \label{#7}
\end{subfigure}
}
\begin{figure}
    \centering
    \buildGraph{100}
    {Loss}{plots/data/femnist-strong-loss}
    {Accuracy}{plots/data/femnist-strong-acc}
    {Favorable network conditions, where $\mathcal{C}_N = 1.0$, $\mathcal{R}_N = 1.0$.}{res:femnist-strong}
    \\[1em]
    \buildGraph{100}
    {Loss}{plots/data/femnist-weak-loss}
    {Accuracy}{plots/data/femnist-weak-acc}
    {Unfavorable network conditions, where $\mathcal{C}_N = 0.5$, $\mathcal{R}_N = 0.5$.}{res:femnist-weak}
    \caption{Training performance for FEMNIST.}
    \label{res:femnist}
\end{figure}
\begin{figure}
    \centering
    \buildGraph{40}
    {Loss}{plots/data/mnist-strong-loss}
    {Accuracy}{plots/data/mnist-strong-acc}
    {Favorable network conditions, where $\mathcal{C}_N = 1.0$, $\mathcal{R}_N = 1.0$.}{res:mnist-strong}
    \\[1em]
    \buildGraph{40}
    {Loss}{plots/data/mnist-weak-loss}
    {Accuracy}{plots/data/mnist-weak-acc}
    {Unfavorable network conditions, where $\mathcal{C}_N = 0.5$, $\mathcal{R}_N = 0.5$.}{res:mnist-weak}
    \caption{Training performance for label-swapped MNIST.}
    \label{res:mnist}
\end{figure}
\begin{figure}
    \centering
    \buildGraph{100}
    {Loss}{plots/data/weatherseq-strong-loss}
    {Loss}{plots/data/weatherseq-strong-loss}
    {Favorable network conditions, where $\mathcal{C}_N = 1.0$, $\mathcal{R}_N = 1.0$.}{res:weatherseq-strong}
    \\[1em]
    \buildGraph{100}
    {Loss}{plots/data/weatherseq-weak-loss}
    {Loss}{plots/data/weatherseq-weak-loss}
    {Unfavorable network conditions, where $\mathcal{C}_N = 0.5$, $\mathcal{R}_N = 0.5$.}{res:weatherseq-weak}
    \caption{Training performance for weather prediction.}
    \label{res:weatherseq}
\end{figure}
\begin{table}[tbhp]
\caption{Scenario Configurations}
\centering
\begin{tabular}{lccccccc}
\hline
\rule{0pt}{9pt}\textsc{dataset} $\mathcal{D}$ & $|N|$ & \textsc{avg.}$|\mathcal{D}_i|$ & \textsc{model} & $T$ & $\beta$ & \textsc{fig.}\\
\hline
FEMNIST & 80 & 108 & LeNet-5 & 100 & 15 & \ref{res:femnist}\\
\shortstack[l]{MNIST \\ {\scriptsize label-swapped}} & 100 & 600 & LeNet-5 & 40 & 3 & \ref{res:mnist} \\
Irish Weather Data & 25 & 3185 & Seq2Seq & 60 & 1 & \ref{res:weatherseq} \\
\hline
\end{tabular}
\label{tab:exp_params}
\end{table}
\begin{table}[htbp]
\centering
\small
\setlength{\tabcolsep}{4pt}
\caption{Final round advantage of \textit{Chisme} vs. baselines. \\
\scriptsize Mean advantages, \emph{$P_{10}$} in parentheses. Acc. entries are absolute percentage-point differences; loss entries are relative improvements. Bolded values denote advantages.}
\label{tab:chisme-vs-baselines}
\begin{tabular}{lcrrrr}
\toprule
Scenario & Metric & FL & FLIC & DFL & GL \\
\midrule
\shortstack[l]{FEMNIST \\ {\scriptsize $\mathcal{C}_N,\mathcal{R}_N=1$}} & \shortstack{acc. \\ {\scriptsize (p.p.)}} & \shortstack{\textbf{+1.48} \\ {\scriptsize \textbf{(+3.29)}}}     & \shortstack{\textbf{+1.44} \\ {\scriptsize \textbf{(+4.06)}}}     & \shortstack{\textbf{+1.47} \\ {\scriptsize \textbf{(+2.89)}}}     & \shortstack{\textbf{+1.59} \\ {\scriptsize \textbf{(+3.35)}}}     \\
\cmidrule(lr){1-6}
\shortstack[l]{FEMNIST \\ {\scriptsize $\mathcal{C}_N,\mathcal{R}_N=1$}} & \shortstack{loss \\ {\scriptsize (\%)}} & \shortstack{\textbf{$-$17.33} \\ {\scriptsize \textbf{($-$28.10)}}} & \shortstack{\textbf{$-$14.56} \\ {\scriptsize \textbf{($-$15.64)}}} & \shortstack{\textbf{$-$17.63} \\ {\scriptsize \textbf{($-$27.87)}}} & \shortstack{\textbf{$-$18.96} \\ {\scriptsize \textbf{($-$33.07)}}} \\
\cmidrule(lr){1-6}
\shortstack[l]{FEMNIST \\ {\scriptsize $\mathcal{C}_N,\mathcal{R}_N=0.5$}} & \shortstack{acc. \\ {\scriptsize (p.p.)}} & \shortstack{\textbf{+1.70} \\ {\scriptsize \textbf{(+3.65)}}}     & \shortstack{\textbf{+1.80} \\ {\scriptsize \textbf{(+2.50)}}}     & \shortstack{\textbf{+1.34} \\ {\scriptsize \textbf{(+2.20)}}}     & \shortstack{\textbf{+1.68} \\ {\scriptsize \textbf{(+3.92)}}}     \\
\cmidrule(lr){1-6}
\shortstack[l]{FEMNIST \\ {\scriptsize $\mathcal{C}_N,\mathcal{R}_N=0.5$}} & \shortstack{loss \\ {\scriptsize (\%)}} & \shortstack{\textbf{$-$19.43} \\ {\scriptsize \textbf{($-$33.35)}}} & \shortstack{\textbf{$-$20.16} \\ {\scriptsize \textbf{($-$34.07)}}} & \shortstack{\textbf{$-$18.40} \\ {\scriptsize \textbf{($-$31.96)}}} & \shortstack{\textbf{$-$20.29} \\ {\scriptsize \textbf{($-$35.58)}}} \\
\cmidrule(lr){1-6}
\shortstack[l]{MNIST (l.s.) \\ {\scriptsize $\mathcal{C}_N,\mathcal{R}_N=1$}} & \shortstack{acc. \\ {\scriptsize (p.p.)}} & \shortstack{\textbf{+4.71} \\ {\scriptsize \textbf{(+9.50)}}}    & \shortstack{$-$1.39 \\ {\scriptsize ($-$2.38)}}                    & \shortstack{\textbf{+4.78} \\ {\scriptsize \textbf{(+9.57)}}}    & \shortstack{\textbf{+10.32} \\ {\scriptsize \textbf{(+15.68)}}}  \\
\cmidrule(lr){1-6}
\shortstack[l]{MNIST (l.s.) \\ {\scriptsize $\mathcal{C}_N,\mathcal{R}_N=1$}} & \shortstack{loss \\ {\scriptsize (\%)}} & \shortstack{\textbf{$-$26.56} \\ {\scriptsize \textbf{($-$30.74)}}} & \shortstack{+17.59 \\ {\scriptsize (+16.24)}}                     & \shortstack{\textbf{$-$26.74} \\ {\scriptsize \textbf{($-$31.08)}}} & \shortstack{\textbf{$-$42.30} \\ {\scriptsize \textbf{($-$47.23)}}} \\
\cmidrule(lr){1-6}
\shortstack[l]{MNIST (l.s.) \\ {\scriptsize $\mathcal{C}_N,\mathcal{R}_N=0.5$}} & \shortstack{acc. \\ {\scriptsize (p.p.)}} & \shortstack{\textbf{+4.95} \\ {\scriptsize \textbf{(+10.29)}}}   & \shortstack{\textbf{+0.89} \\ {\scriptsize \textbf{(+2.57)}}}    & \shortstack{\textbf{+4.33} \\ {\scriptsize \textbf{(+9.09)}}}    & \shortstack{\textbf{+8.44} \\ {\scriptsize \textbf{(+13.69)}}}   \\
\cmidrule(lr){1-6}
\shortstack[l]{MNIST (l.s.) \\ {\scriptsize $\mathcal{C}_N,\mathcal{R}_N=0.5$}} & \shortstack{loss \\ {\scriptsize (\%)}} & \shortstack{\textbf{$-$28.08} \\ {\scriptsize \textbf{($-$33.82)}}} & \shortstack{\textbf{$-$8.08} \\ {\scriptsize \textbf{($-$3.99)}}} & \shortstack{\textbf{$-$26.24} \\ {\scriptsize \textbf{($-$32.76)}}} & \shortstack{\textbf{$-$39.38} \\ {\scriptsize \textbf{($-$47.16)}}} \\
\cmidrule(lr){1-6}
\shortstack[l]{Weather Pred. \\ {\scriptsize $\mathcal{C}_N,\mathcal{R}_N=1$}} & \shortstack{loss \\ {\scriptsize (\%)}} & \shortstack{\textbf{$-$1.71} \\ {\scriptsize \textbf{($-$1.37)}}}  & \shortstack{\textbf{$-$3.16} \\ {\scriptsize \textbf{($-$3.29)}}}  & \shortstack{\textbf{$-$1.65} \\ {\scriptsize \textbf{($-$1.58)}}}  & \shortstack{\textbf{$-$2.99} \\ {\scriptsize \textbf{($-$2.69)}}}  \\
\cmidrule(lr){1-6}
\shortstack[l]{Weather Pred. \\ {\scriptsize $\mathcal{C}_N,\mathcal{R}_N=0.5$}} & \shortstack{loss \\ {\scriptsize (\%)}} & \shortstack{\textbf{$-$2.00} \\ {\scriptsize \textbf{($-$2.04)}}}  & \shortstack{\textbf{$-$7.30} \\ {\scriptsize \textbf{($-$6.82)}}}  & \shortstack{\textbf{$-$2.08} \\ {\scriptsize \textbf{($-$1.95)}}}  & \shortstack{\textbf{$-$3.48} \\ {\scriptsize \textbf{($-$3.16)}}}  \\
\bottomrule
\end{tabular}
\end{table}

\subsection{Methodology}
We test \textit{Chisme} against standard FL, DFL, and GL implementations as described in Section~\ref{sec:bg-related}.
In addition, we include FLIC~\cite{castellon2022flic} in our comparative study 
to represent approaches that, similar to \textit{Chisme}, leverage cosine similarity to cluster clients for model aggregation, setting FLIC's clustering round to occur only after the sufficient rounds of training necessary for effective clustering. 

To evaluate \textit{Chisme}'s performance, we use two different use cases, namely, image recognition and weather prediction.
As a representative of image recognition scenarios, we use the FEMNIST~\cite{caldas2018leaf} dataset which represents a real-world, distributed version of MNIST -- a standard handwritten digit dataset -- with a natural level of incongruence between clients' data distributions, where each client's dataset contains the samples of a different real-world client.
To observe performance in extreme heterogeneity, we also include a modified MNIST dataset in which we introduce artificial heterogeneity among clients' data via \textit{label-swapping} (as done in other heterogeneity-focused FL experiments~\cite{sattler2020cfl, castellon2022flic}), in which 5 different data distribution groups swap different pairs of labels to induce incongruence between clients' datasets.
To represent the sequential regression use case, we used the Irish weather station dataset which was sourced from the Irish meteorological service \textit{Met Éiranne}~\cite{meteireann_climate_data}, which contains entries for a single year across 25 weather stations, using temperature, humidity, rain, and dew-point as input and prediction features.
In all scenarios, dataset sizes among clients vary.
Number of clients ($|N|$), average sample count per client (\textsc{avg.}$|\mathcal{D}|$), model, number of communication rounds per client ($T$), and number of training epochs ($\beta$) are shown in Table~\ref{tab:exp_params}. 
Deep learning models (\textit{LeNet-5}~\cite{lecun1998gradient} and \textit{Seq2Seq}~\cite{xu2024multi}) were implemented in the PyTorch framework.
Within each scenario and configuration, all paradigms have identical communication budgets, i.e., exchange the same number of messages over the network.
We use a Watts-Strogatz model~\cite{watts1998collective} to represent real-world peer-to-peer network topologies.
We configure two network parameters -- \textit{connectivity} and \textit{reliability}.
Connectivity ($\mathcal{C}_N$) refers to how connected each node is to all other nodes, where $\mathcal{C}_N = 1$ denotes a fully connected network and $\mathcal{C}_N = 0$ denotes a minimally connected ring topology.
Reliability ($\mathcal{R}_N$) refers to how reliable the connections between nodes are when communicating over them, where $\mathcal{R}_N = 1$ denotes no loss over the connections and $\mathcal{R}_N = 0$ denotes complete loss over the network.
Clients were evaluated on their own dataset at every round, after training such that the model parameters could be fine-tuned locally after aggregation or merging.

\subsection{Results}
Results averaged across multiple seeds are shown in Figure~\ref{res:femnist} (FEMNIST), Figure~\ref{res:mnist} (MNIST), and Figure~\ref{res:weatherseq} (Irish Weather Station).
We observe performance through mean loss and/or accuracy across clients and observe the $P_{10}$ quantile to infer performance equity among clients.
Comparisons are shown in Table~\ref{tab:chisme-vs-baselines}.
However, overall, we observe that \textit{Chisme} outperforms baselines in almost every scenario, achieving higher accuracy, lower loss, and faster convergence -- by at least $13\%$ (FEMNIST, $\mathcal{C}_N,\mathcal{R}_N=1$) and up to $36\%$ (MNIST, $\mathcal{C}_N,\mathcal{R}_N=0.5$) fewer rounds to reach $90\%$ final performance when compared to the next best approach -- while exhibiting less performance disparity between clients.
These advantages are more pronounced under weaker network conditions, in which \textit{Chisme} exhibits relatively little performance degradation.
The notable outlier is with the artificially incongruent MNIST dataset, in which FLIC has a slight advantage in the stronger $\mathcal{C}_N,\mathcal{R}_N=1$ network scenario, but \textit{Chisme} retains its advantage in the weaker $\mathcal{C}_N,\mathcal{R}_N=0.5$ network scenario.
Given the strong advantage \textit{Chisme} provides for the $P_{10}$ quantile over other approaches, we infer that leveraging our gossip-adapted cosine similarity-based heuristic enables GL's merging to accelerate training, such that struggling clients can speed up model progression by merging in models from more capable clients with similar data distributions and therefore similar but more experienced model progressions.
Our observations also suggest that, as clients train and collaborate, \textit{Chisme} pushes clients to graduate transition from broad to selective collaboration, as opposed to discrete stages of broad then selective collaboration~\cite{huang2021personalized, arivazhagan2019silopersonalized, sattler2020cfl}.
\section{Conclusion} \label{sec:conclusion}

In this paper, we introduced \textit{Chisme}, a fully decentralized and heterogeneity-aware approach to distributed deep learning.
Our evaluation of \textit{Chisme} demonstrates better and more equitable performance outcomes than existing approaches for all clients despite heterogeneity in dataset sizes, experience levels, and data distributions, with pronounced advantage in sparser, more volatile networks.
Practical viability of \textit{Chisme} and similar approaches also hinges on future system-level evaluation that will evaluate beyond loss and convergence to include network bandwidth utilization, energy consumption profiles, and CPU and I/O utilization.
As end users are exposed to a more diverse selection of models operating on more diverse networks, we also aim to explore how we can extend our methods to achieve collaboration in the face of more heterogeneity, e.g., diverse model architectures and task alignments.
The work presented in this paper along with its research trajectory contribute to the emerging class of robust protocols that can enrich and empower an increasingly diverse and distributed data landscape.
\bibliographystyle{IEEEtran}
\bibliography{references.bib}
\end{document}